\documentclass[lettersize,journal]{IEEEtran}
\usepackage{amsmath,amsfonts}
\usepackage{algorithmic}
\usepackage{algorithm}
\usepackage{array}
\usepackage[caption=false,font=normalsize,labelfont=sf,textfont=sf]{subfig}
\usepackage{textcomp}
\usepackage{stfloats}
\usepackage{url}
\usepackage{verbatim}
\usepackage{graphicx}
\usepackage{gensymb}
\usepackage{cite}
\usepackage{hyperref}
\usepackage{xspace}

\newcommand{\ourData}{\textit{EMG-FK}\xspace}
\newcommand{\ourModel}{\textit{TRR}\xspace}

\begin{document}

\title{Decoding High-Dimensional Finger Motion from EMG Using Riemannian Features and RNNs}

\author{
\IEEEauthorblockN{
Martin Colot\IEEEauthorrefmark{1}\IEEEauthorrefmark{4},
Cédric Simar\IEEEauthorrefmark{1}\IEEEauthorrefmark{4},
Guy Cheron\IEEEauthorrefmark{2},
Ana Maria Cebolla Alvarez\IEEEauthorrefmark{2},
Gianluca Bontempi\IEEEauthorrefmark{1}\IEEEauthorrefmark{3}
}

\IEEEauthorblockA{\IEEEauthorrefmark{1}
Machine Learning Group, ULB, Bruxelles, Belgium}

\IEEEauthorblockA{\IEEEauthorrefmark{2}
Neuromove, ULB Neuroscience Institute, ULB, Brussels, Belgium}

\IEEEauthorblockA{\IEEEauthorrefmark{3}
WEL Research Institute, Wavre, Belgium}

\IEEEauthorblockA{\IEEEauthorrefmark{4}
These authors contributed equally to this work and share first authorship.}

\IEEEauthorblockA{
\texttt{martin.colot@ulb.be}
}
}

\maketitle              

\begin{abstract}
Continuous estimation of high-dimensional finger kinematics from forearm surface electromyography (EMG) could enable natural control for hand prostheses, AR/XR interfaces, and teleoperation. However, the complexity of human hand gestures and the entanglement of forearm muscles make accurate recognition intrinsically challenging. Existing approaches typically reduce task complexity by relying on classification-based machine learning, limiting the controllable degrees of freedom and compromising on natural interaction. We present an end-to-end framework for continuous EMG-to-kinematics regression using only consumer-grade hardware. The framework combines an 8-channel EMG armband, a single webcam, and an automatic synchronization procedure, enabling the collection of the \textit{EMG Finger-Kinematics} dataset (\ourData), a 10-h dataset of synchronized EMG and 15 finger joint angles from 20 participants performing rich, unconstrained right-hand motions. We also introduce the \textit{Temporal Riemannian Regressor} (\ourModel), a lightweight GRU-based model that uses sequences of multi-band Riemannian covariance features to decode finger motion. Across \ourData\ and the public \textit{emg2pose} benchmark, \ourModel outperforms state-of-the-art methods in both intra- and cross-subject evaluation. On \ourData, it reaches an average absolute error of $9.79\degree \pm 1.48$ in intra-subject and $16.71 \degree \pm 3.97$ in cross-subject. Finally, we demonstrate real-time deployment on a Raspberry Pi 5 and intuitive control of a robotic hand; \ourModel\ runs at nearly 10 predictions/s and is roughly an order of magnitude faster than state-of-the-art approaches. Together, these contributions lower the barrier to reproducible, real-time EMG-based decoding of high-dimensional finger motion, and pave the way toward more natural and intuitive control of embedded EMG-based systems.
\end{abstract}


\section{Introduction~\label{sec:intro}}
Emerging paradigms in human–computer interaction are moving beyond traditional device-centric mechanisms such as mice, keyboards, buttons, and joysticks, which require users to conform to artificial control schemes. Instead, modern paradigms favor more integrated, natural, and unconstrained interaction modalities. Examples of such modalities include wearable technologies (e.g., smartwatches and augmented reality glasses) and behavioral sensing (e.g., eye tracking, voice recognition, facial expression analysis, and full-body motion tracking). Collectively, these approaches aim to reduce operational burden and reposition digital technology from a tool that must be explicitly operated to an intelligent and adaptive system.
Hand gesture recognition (HGR) is a particularly compelling technology, as it allows users to control a piece of software without manipulating a physical input device and thus remain available for concurrent physical activities.
HGR is commonly implemented using cameras and computer vision to track joint positions~\cite{zhang2020mediapipe}~\cite{spurr2021self}. While this can achieve rich motion capture, the performance often degrades under hand occlusion, poor lighting conditions, or during object manipulation~\cite{buckingham2021hand}. Additionally, these systems depend on a camera positioned in the environment or mounted on the user, which may constrain mobility, increase setup complexity, and raise privacy concerns.

A promising alternative is electromyography (EMG)-based HGR. In this approach, users are equipped with physiological sensors positioned around the forearm or the wrist to measure their muscle activity, which can be processed to infer hand and finger movements. EMG-based interaction reduces dependence on external sensing infrastructure and is less affected by lighting conditions or occlusion, making it a minimally intrusive and robust solution for hands-free input.
This technology is gaining increasing attention for applications in augmented and extended reality (AR/XR)~\cite{toledo2022virtual}, teleoperation~\cite{hassan2020teleoperated}, rehabilitation~\cite{yun2017maestro}, and sign language recognition~\cite{ben2023sign}, where minimally invasive and low-latency input modalities are essential. EMG sensing also plays a significant role in the control of robotic hand prostheses~\cite{chen2023review}. Because most of the muscles responsible for finger movements are located in the forearm, individuals with upper-limb loss can voluntarily contract residual muscles, allowing EMG systems to detect muscle activation patterns and infer intended finger motions for prosthetic control~\cite{unanyan2021design}.

Hand gesture recognition from EMG relies heavily on machine learning~\cite{ting2022review}~\cite{tamilvanan2025comprehensive}. Large quantities of EMG signals are collected and synchronized with related kinematic information. Predictive models are then trained to infer hand kinematics from recent EMG observations. Because EMG signals are highly sensitive to individual differences (e.g., muscle size, gesture patterns, and mental state), sensor brand and placement, and the set of target gestures, each EMG-based system typically requires a dedicated data acquisition phase to develop a customized model that accurately reflects the needs of the application.
Due to the complexity of the task, EMG-based applications generally rely on classification models trained to recognize a discrete set of hand postures~~\cite{simar2024machine}~\cite{tamilvanan2025comprehensive}. As a result, the predicted kinematics change instantaneously when the user transitions from one posture to another.
These models have demonstrated high accuracy in \textit{intra-subject} configurations~\cite{simar2024machine}. This setting minimizes the effects of shifts in EMG distributions due to individual session characteristics by retraining the machine learning model with data collected from the same person within the same session.
Recent studies~\cite{colot2026linear}~\cite{metaEMG2pose}~\cite{kaifosh2025generic} have focused on improving robustness across new subjects and sessions, thereby enhancing the accessibility of these models by reducing the duration of calibration procedures.
However, classification-based systems often feel too restrictive and fail to provide a natural user experience. In particular, many users of myoelectric hand prostheses have reported this limitation as one of the reasons that led them to abandon the device~\cite{salminger2022current}.
A more suitable approach is to develop regression models capable of continuously and simultaneously estimating the joint-angles of a hand with multiple degrees of freedom (DOF)~\cite{metaEMG2pose}. Such models impose no constraints on combinations of joint configurations or velocities, enabling smooth and natural hand movement reconstruction and thereby enhancing the perception of natural use.

The regression problem is particularly challenging due to the large number of degrees of freedom of the human hand~\cite{elkoura2003handrix}. Existing works addressing this topic share several common characteristics.
The first is that studies often introduce constraints during data collection by restricting participant gestures, thereby reducing the dimensionality of hand kinematics. Participants are instructed to perform guided postures selected from a predefined and limited set~\cite{simpetru2022accurate}~\cite{quivira2018translating}~\cite{krasoulis2019effect}~\cite{tacca2024wearable}~\cite{dwivedi2020emg}~\cite{stapornchaisit2019finger}~\cite{qin2021multi}~\cite{hai2022simultaneous}~\cite{koch2021semg}~\cite{avian2022estimating}. However, this approach ultimately restricts the distribution of joint-angles to discrete combinations, effectively reducing the problem to simpler classification and potentially leading to overly optimistic performance estimates.
Alternatively, in some other studies, participants are instructed to follow predefined trajectories at imposed speeds (e.g., sinusoidal finger motions)~\cite{simpetru2022accurate}~\cite{hai2022simultaneous}. While this approach requires the development of regression models capable of estimating joint-angles in real time, the resulting complexity of gesture distributions remains unrealistic with respect to real-world applications.
Another characteristic of the existing works is the limited diversity of model architectures employed in joint-angle regression. For feature extraction, studies have primarily relied on either time-domain features (TDF)~\cite{quivira2018translating}~\cite{tacca2024wearable, dwivedi2020emg}~\cite{stapornchaisit2019finger}~\cite{hai2022simultaneous}~\cite{avian2022estimating} or convolutional neural networks (CNN)~\cite{metaEMG2pose}~\cite{simpetru2022accurate}~\cite{liu2021neuropose}~\cite{simpetru2022sensing}~\cite{qin2021multi}. 
Convolutions are applied along the time axis (1D CNN) for individual EMG channels or across both spatial (EMG grid) and temporal dimensions (3D CNN) for high-density EMG sensors~\cite{simpetru2022accurate}.
TDF are simple to compute and practical when training data is limited, but they often yield suboptimal regression performance. In contrast, CNN combined with deep neural networks can achieve improved results but typically require more training data. Few studies explore alternative feature representations, such as frequency-domain features~\cite{avian2022estimating}, motor unit action potentials (MUAPs)~\cite{simpetru2022sensing}, or raw EMG signals without convolution~\cite{koch2021semg}.
Regarding regression algorithms, basic approaches typically employ relatively deep dense neural networks~\cite{simpetru2022sensing}~\cite{tacca2024wearable}~\cite{qin2021multi}~\cite{hai2022simultaneous}. Moreover, because EMG signals reflect movement dynamics rather than absolute joint configurations, the activation pattern associated with a given hand configuration depends on the preceding movement history.
This naturally motivates the use of recurrent models, such as long short-term memory networks (LSTMs), gated recurrent unit (GRU), or transformer architectures to predict continuous sequences of joint-angles based on extended EMG sequences, thereby providing temporal context \cite{metaEMG2pose}~\cite{liu2021neuropose}~\cite{quivira2018translating}~\cite{sosin2018continuous}~\cite{koch2021semg}~\cite{avian2022estimating}~\cite{ngeo2014}~\cite{bao2021}~\cite{zanghieri2021}~\cite{putro2024}~\cite{lin2024}~\cite{lin2025}~\cite{metaEMG2pose}.
This paper will study the impact of covariance-based methods using Riemannian geometry. They have been explored in biosignal processing as an alternative feature extraction approach~\cite{BARACHANT2013_KERNEL}~\cite{SIMAR2020_FVMSD}~\cite{eyvazpour2026}. However, their application to kinematics regression remains limited. Within the sEMG domain, Jaquier and Calinon demonstrated the potential of geometry-aware covariance modeling for continuous decoding of wrist kinematics ~\cite{jaquier2017}~\cite{jaquier2017_emg_tmg}. Additionally, covariance-based features manipulated with Riemannian geometry have proven effective for decoding finger trajectories from other modalities such as ECoG~\cite{yao2022}~\cite{eyvazpour2026}, highlighting the utility of geometry-aware covariance descriptors in complex, high-dimensional kinematic regression tasks. Building upon these insights, we introduce, to our knowledge, the first approach integrating multi-band covariance matrices projected from the Riemannian manifold to the Euclidean tangent space (CMTS) with a recurrent neural network for finger joint-angle regression from sEMG, achieving high accuracy alongside suitability for embedded, real-time inference.

A recent study by \textit{Meta Reality Labs}~\footnote{Meta Reality Labs: \url{https://tech.facebook.com/reality-labs/}}~\cite{metaEMG2pose} represents a significant advancement in this area. The authors collected a large dataset comprising 193 participants across multiple sessions, performing both guided and free-hand gestures. This work also proposes a deep recurrent regression neural network, called \textit{vemg2pose}, that combines CNN and LSTM layers. This model predicts joint velocities, which are then integrated to obtain hand kinematics. While their contribution is substantial, replication is challenging due to the complexity of the data acquisition protocol, which relies on multiple cameras and custom EMG sensors. 

In this paper, we aim to lower the barrier to continuous EMG-to-kinematics regression for HGR by proposing an accessible, novel data acquisition framework and introducing a novel regression model optimized for real-time deployment, outperforming state-of-the-art baselines on our dataset and the \textit{emg2pose} benchmark~\cite{metaEMG2pose}.
The main original contributions of this paper are: 
(i) a reproducible data acquisition framework enabling high-quality EMG-kinematics recordings using consumer-grade equipment, and a novel synchronization method, specifically designed to be cost-effective and deployable outside laboratory conditions to support ongoing calibration and model refinement. This framework leverages minimal hardware, requiring only an 8-channel EMG armband (\textit{MindRove}~\footnote{MindRove EMG armband 8 channels: \url{https://mindrove.com/product/emg-armband/}}) and a standard laptop camera, to provide synchronized sEMG and hand-kinematics data streams;
(ii) an original dataset, called \textit{EMG-Finger-Kinematics} (\ourData)~\cite{EMGFK_dataset}, of unconstrained right-hand gestures from 20 participants (30 minutes per participant); 
(iii) a systematic benchmark of state-of-the-art EMG-to-kinematics regressors, covering both feature-engineering and representation-learning approaches, on both our \ourData dataset and the public \textit{emg2pose} benchmark~\cite{metaEMG2pose};
(iv) a novel lightweight regression approach, called \textit{Temporal Riemannian Regressor} (\ourModel), combining multi-band covariance-matrix features projected from the Riemannian manifold to an Euclidean tangent space with a recurrent neural network, demonstrating more accurate estimations of high-degree-of-freedom finger joint angles from surface EMG compared to state-of-the-art methods; 
(v) an embedded deployment study demonstrating the proposed \ourModel model's capability for real-time inference on compact edge devices, while respecting practical thermal constraints, an essential consideration for real-world prosthetic control.

\section{Methods}

\subsection{Data acquisition setup}

\begin{figure*}
    \centering
    \includegraphics[width=1\linewidth]{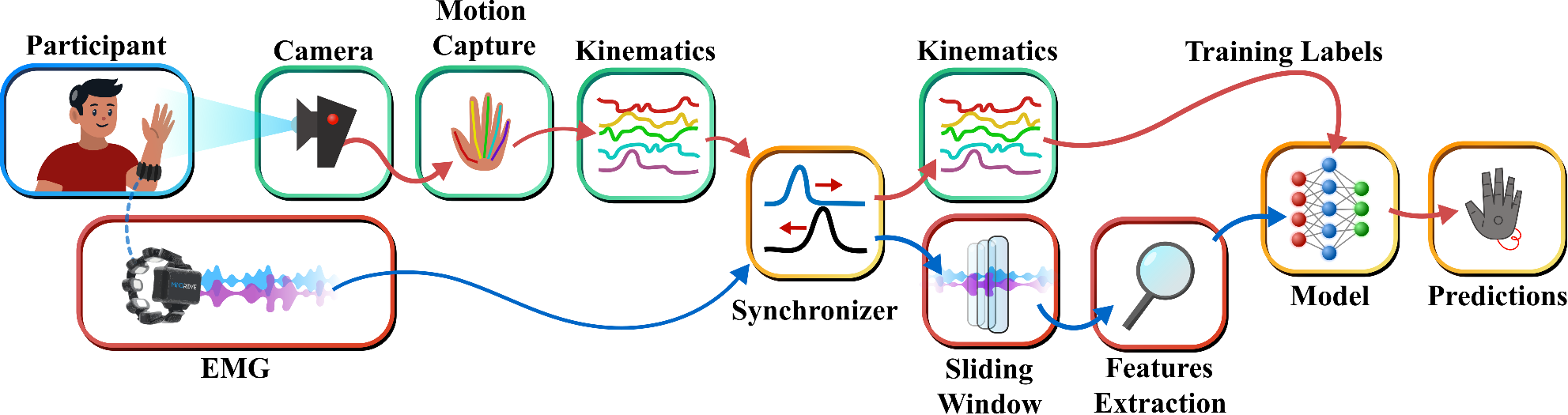}
    \caption{Pipeline for data acquisition and processing}
    \label{fig:dataAcquisition}
\end{figure*}

The data acquisition setup is illustrated in Figure~\ref{fig:dataAcquisition} and consists of two main modules: EMG acquisition and motion capture.
EMG data are collected using the \textit{MindRove 8-channel EMG armband}.
It features eight dry electrodes evenly positioned around the forearm to record eight EMG channels at a sampling frequency of 500 Hz. The signals are transmitted via a dedicated Wi-Fi connection to a consumer-grade laptop.
This EMG device is increasingly used as an alternative to medical-grade EMG systems in research contexts that prioritize more natural usage conditions, broader accessibility at lower cost, and reduced power consumption~\cite{taori2024use}~\cite{taori2025classification}. These characteristics make it better suited for embedded systems compared to high-density EMG sensors.

To acquire joint-angle kinematics, we use a standard laptop webcam and the \textit{MediaPipe} package~\cite{zhang2020mediapipe}. This software relies on computer vision to estimate anatomical landmarks on the fingers. Relative angles between triplets of consecutive joints are computed to obtain a measure of joint flexion, which can be linearly mapped to realistic joint-angle ranges. 
The kinematic signal is then resampled to 500 Hz using linear interpolation to match the EMG sampling frequency.
This camera-based framework is low-cost compared to multiple camera setups or cyber gloves, yet it provides a sufficient quality of motion capture and can be ran at home by the user, enabling post-calibration without expensive equipment.

We obtain 15 channels of kinematic data and 8 channels of EMG. These signals are then synchronized using the method described in Section~\ref{sec:sync} and processed with sliding windows.

\subsection{Data acquisition protocol~\label{sec:acquisitionProtocol}}

The subject is first equipped with the recording devices.
The EMG armband is positioned at 20 cm from the wrist, with the main module placed on the top side of the forearm when the hand is in pronation. Particular attention is paid to ensure proper skin contact to maintain a high signal-to-noise ratio. The EMG signals are then filtered with a band-pass filter (15-150 Hz) to remove artifacts and a notch filter at 50 and 100 Hz to remove power-line artifacts.
The subject is then seated in front of the laptop and webcam, with the elbow comfortably resting on a table. The palm is oriented toward the webcam throughout the acquisition.
During data collection, participants are instructed to perform free and unconstrained finger gestures for 30 minutes. Each joint articulation is treated as having a single DOF, meaning that fingers are allowed to open and close but not to cross or apply force on each other. 
This ensures that the recorded EMG activity primarily reflects dynamic muscle contractions rather than sustained isometric activation.
No restrictions are imposed on movement velocities, joint-angle configurations, or gesture sequences.
During the first 15 minutes, the participant sees on the screen a virtual robotic hand (representation of the \textit{Leap Hand}~\cite{shaw2023leap}) which follows the camera-based motion capture. This visual feedback helps the subject understand the tracking capabilities and produce gestures within the range reliably captured by the vision system.
A short break of approximately 5 minutes follows. During this time, the acquisition software uses the previously recorded data to train an EMG-based regression model (as described in Section~\ref{sec:ourModel}). In the final 15 minutes, the participant continues performing free gestures, but the virtual hand feedback is now controlled by the EMG-based model's predictions.
The rationale behind this protocol, inspired by active learning and human in the loop interaction~\cite{nawfel2022influence}~\cite{hahne2015concurrent}, is to allow participants to naturally explore joint configurations and movement speeds that are not well predicted by the model, thereby enriching the dataset and improving coverage of the expected motion space.

We recorded data with 20 volunteer participants, using their right hand, resulting in the 10-hour-long \ourData dataset.

\subsection{Synchronization~\label{sec:sync}}

Synchronization between EMG and kinematics is a critical and often overlooked issue~\cite{schulte2022synchronization}.
Because no dedicated hardware is used to synchronize EMG and kinematic recordings, a small temporal offset between the two signals may exist. However, this offset remains constant throughout the acquisition, as both signals are captured within a single continuous loop, the EMG sampling rate is precisely 500 Hz, and relative timestamps are available for each kinematic frame. These factors ensure that any misalignment is primarily due to fixed system latency and can therefore be corrected by applying a single consistent time shift.

\begin{figure*}
    \centering
    \includegraphics[width=1\linewidth]{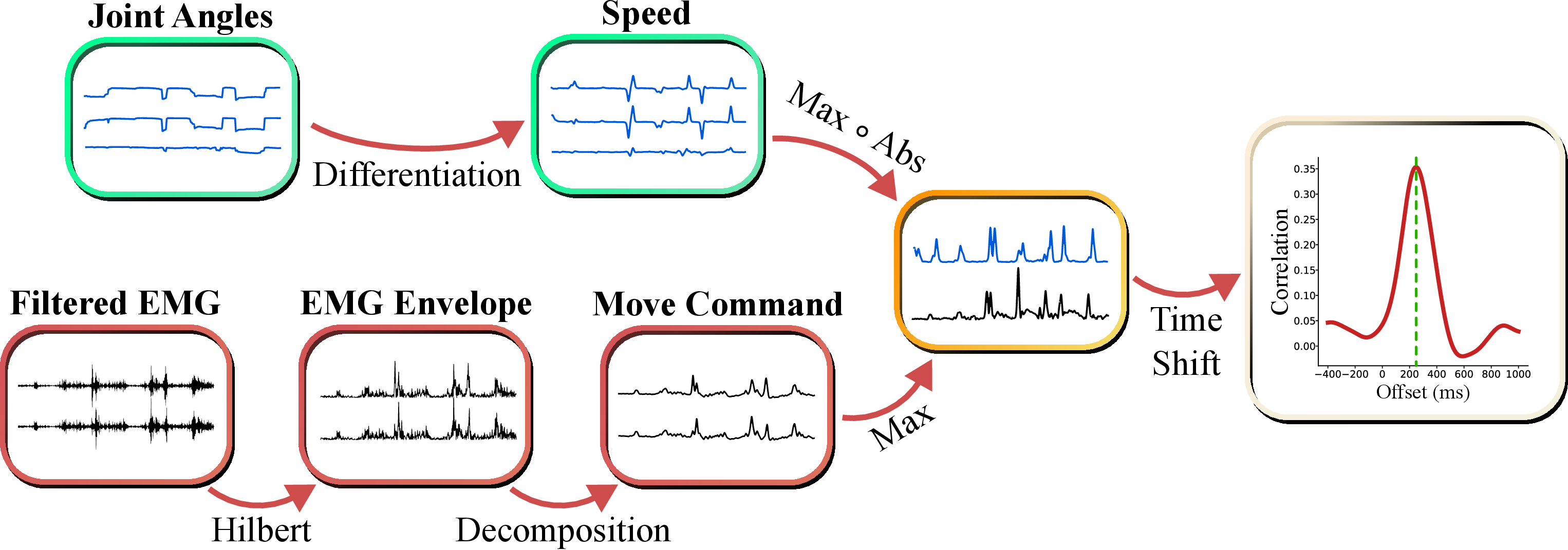}
    \caption{Automatic procedure for synchronization of EMG and joint-angles}
    \label{fig:synchro}
\end{figure*}

To identify this time shift, we developed an automatic synchronization procedure. The method is based on the idea that EMG signals can be decomposed into \textit{move} and \textit{hold} commands using Hilbert transform and an integration approach~\cite{simar2024machine}, and that the amplitude of the \textit{move} command directly determines the speed of the subsequent gestures.
Our synchronization method, illustrated in Figure~\ref{fig:synchro}, determines the temporal shift that maximizes the correlation between the maximum absolute velocity across all joints and the maximum \textit{move} command across all EMG channels at each time frame.
This synchronization process is evaluated in Appendix~\ref{sec:evalSync}.
To ensure proper synchronization, this procedure was applied independently to the first and second halves of each participant’s recordings. 

\subsection{The \ourModel model}

Our method, \ourModel, builds on insights gained from prior related work on hand gesture classification~\cite{simar2024machine}~\cite{colot2026linear}. In particular, covariance matrices projected into the tangent space have proven to be highly effective feature representations, reducing the need for complex deep neural architectures. Additionally, inter-subject variability in EMG signals motivates the use of subject-specific models. So, we designed a relatively shallow architecture based on CMTS features and optimized it for intra-subject performance.

Moreover, we use the observation that EMG activity is context-dependent and more closely related to movement velocity than to static posture. In other words, the EMG produced when reaching a given joint-angles configuration depends on the preceding posture. This motivates the use of a recurrent model. Furthermore, the task is inherently multivariate, and nonlinear relationships may exist between different joint-angles. Therefore, a nonlinear model such as a multilayer neural network is well-suited.
We experimented with a range of models, from traditional methods (ridge regression, gradient boosting) to deeper neural architectures, while evaluating various feature extraction techniques and input sequence configurations. These analyses led us to propose a model that combines sequences of CMTS feature vectors extracted from short EMG epochs with a lightweight recurrent neural network whose limited size enables real-time, on-device inference.
All the model's hyperparameters were determined through an ablation study (presented in section~\ref{sec:ablation_study}) to optimize the predictions on our dataset. The resulting configuration was then validated by applying the same model and hyperparameters to the \textit{emg2pose} dataset.

\subsubsection{Riemannian feature extraction}
Covariance matrices and their subsequent projection from the Riemannian manifold to an Euclidean tangent space have recently demonstrated superiority over standard features for classification-based hand gesture recognition from EMG~\cite{manjunatha2020classification}~\cite{simar2024machine}~\cite{colot2026linear}. These features are known to provide a rich representation of brain activation in EEG recordings~\cite{lotte2018review}, as they capture both independent point activations and connectivity between different brain regions. In the context of EMG processing, covariance matrices encode patterns of individual and joint muscle activations, which appear to be more informative than channel-wise TDF features or CNN-based representations. We compute CMTS for individual samples following the procedure described in~\cite{simar2024machine}.
For each sample, we extract 10 EMG windows of 300 ms with a step size of 100 ms. For each window, CMTS features are computed in three frequency bands (5–40 Hz, 40–80 Hz, and 80–150 Hz), resulting in a sequence of concatenated feature vectors. We also propose a simplified version of \ourModel where CMTS are computed in a single full frequency band (5-150 Hz).

\subsubsection{Model architecture~\label{sec:ourModel}}
\ourModel is a recurrent neural network based on GRU units, using the engineered features described below as input. 
The network begins with a dense layer of 256 units with tanh activation, followed by two stacked GRU layers with 256 and 128 units, respectively, both using tanh activation and a dropout rate of 10\%. A final hidden dense layer of 64 units with tanh activation is then applied.
Training is performed using the Adam optimizer with Huber loss, a clip norm of 1, a learning rate of $2 \times 10^{-4}$, and a batch size of 256. Early stopping with a patience of 20 epochs is used to determine convergence.

\subsection{Benchmark models}

\subsubsection{vemg2pose}
We use the public Python implementation of \textit{vemg2pose} available on GitHub~\footnote{vemg2pose: \url{https://github.com/facebookresearch/emg2pose}}. The training parameters are set to the values reported as optimal in \cite{metaEMG2pose}, including a batch size of 64 and a learning rate of $10^{-3}$.
Also, since this model expects EMG signals sampled at 2000 Hz as input, the signals from our dataset are resampled using linear interpolation to match this sampling rate.

\subsubsection{Time Domain Features (TDF)~\label{sec:TDF}}
We compute the following features for each EMG channel: mean absolute value, root mean square, maximum absolute amplitude, waveform length, slope sign change, Wilson amplitude, and maximum fractal length \cite{simar2024machine}. The samples are then standardized using the mean and standard deviation computed from the training set, and subsequently used as input to a multilayer perceptron (MLP) regressor.

\subsubsection{Multi layers perceptrons (MLP)~\label{sec:mlp}}
We use standard MLPs to perform regression from either TDF features or CMTS. MLPs are widely used in machine learning applications involving EMG~\cite{ting2022review} and EEG~\cite{xie2020review}. The network architecture consists of four fully connected hidden layers with 256, 256, 128, and 64 units, respectively, all using tanh activation. Training is performed with the Adam optimizer and Huber loss, using a learning rate of $2 \times 10^{-4}$, a clip norm of 1, and a batch size of 512. Early stopping with a patience of 20 epochs is applied.

\subsubsection{Deep recurrent and convolutional neural network (CRNN)~\label{sec:deepCRNN}}
We apply five consecutive one-dimensional convolutional layers along the time axis, each with 32 filters, a kernel size of 3, and ReLU activation. Max pooling with a pool size of 2 is performed after the first four convolutional layers. Batch normalization is then applied, and the signal is segmented into windows of length 20 with a step size of 6.
Next, a time-distributed dense layer with 128 units and ReLU activation is used, followed by two stacked GRU layers with 128 and 64 units, respectively, both using tanh activation. A final hidden dense layer of size 64 with tanh activation is then applied.
The model is trained using the Adam optimizer with Huber loss, a learning rate of $2 \times 10^{-4}$, a batch size of 512, and early stopping with a patience of 20 epochs.

\subsection{Evaluation}
We use two approaches for evaluation.
The first is intra-subject, single-session. This configuration minimizes the domain discrepancy~\cite{colot2026linear} arising from subject-specific distributions of EMG and finger kinematics.
For each subject and model, we perform 10-fold cross-validation without shuffling the samples to prevent data leakage. Before training, a random continuous segment of signal representing 10\% of the training set is selected as the validation set. The model is then trained on the remaining training data until convergence on the validation set. Predictions from the test sets across all folds are concatenated to obtain a complete performance evaluation for each subject. 
The second evaluation approach is cross-subject, single-session. This configuration is expected to yield sub-optimal predictions for each test subject. However, it emulates a more convenient system that does not require calibration for the final user.
This evaluation is performed using leave-one-subject-out (LOSO) cross-validation. With $N$ subjects, we compute $N$ folds, where one subject is used for testing, one is chosen at random for validation, and the remaining $N-2$ for training. Also, to reduce person-specific variability, we standardize the EMG signals individually for each subject. This reduces the effect of muscle size but does not eliminate the influence of the other factors.

From a mathematical point of view, the intra-subject configuration helps to estimate the distribution $P(\text{Kinematic} \mid \text{EMG}, \lambda_S)$, where $\lambda_S$ is the unknown variable influencing the distribution of kinematic and EMG from subject $S$. The cross-subject configuration enables the estimation of the more general distribution $P(\text{Kinematic} \mid \text{EMG})$.
Together, they define an upper bound (intra-subject) and a lower bound (cross-subject) on the performance that models can achieve.

To emulate real-time continuous prediction, samples are extracted using a sliding window with a 100 ms step size and a window length determined by the model.
The evaluation is carried out on both the \ourData and \textit{emg2pose}~\cite{metaEMG2pose} datasets. From the latter, we select the first large session\footnote{Marked as \textit{demonstration} session in the dataset folder} of each of the first 30 subjects. For each session, we concatenate the first halves of all recordings, followed by the second halves. Because different recordings within a session correspond to different gesture exercises, this reorganization ensures that the test data in each fold includes samples from recordings that are also represented in the training set. This reduces kinematic discrepancy between training and testing while still minimizing data leakage.
We use the average normalized mean square error (NMSE) across all joint-angles as the evaluation metric.

\section{Results and discussions}

\subsection{Data analysis} 

\begin{figure*}
    \centering
    \includegraphics[width=1\linewidth]{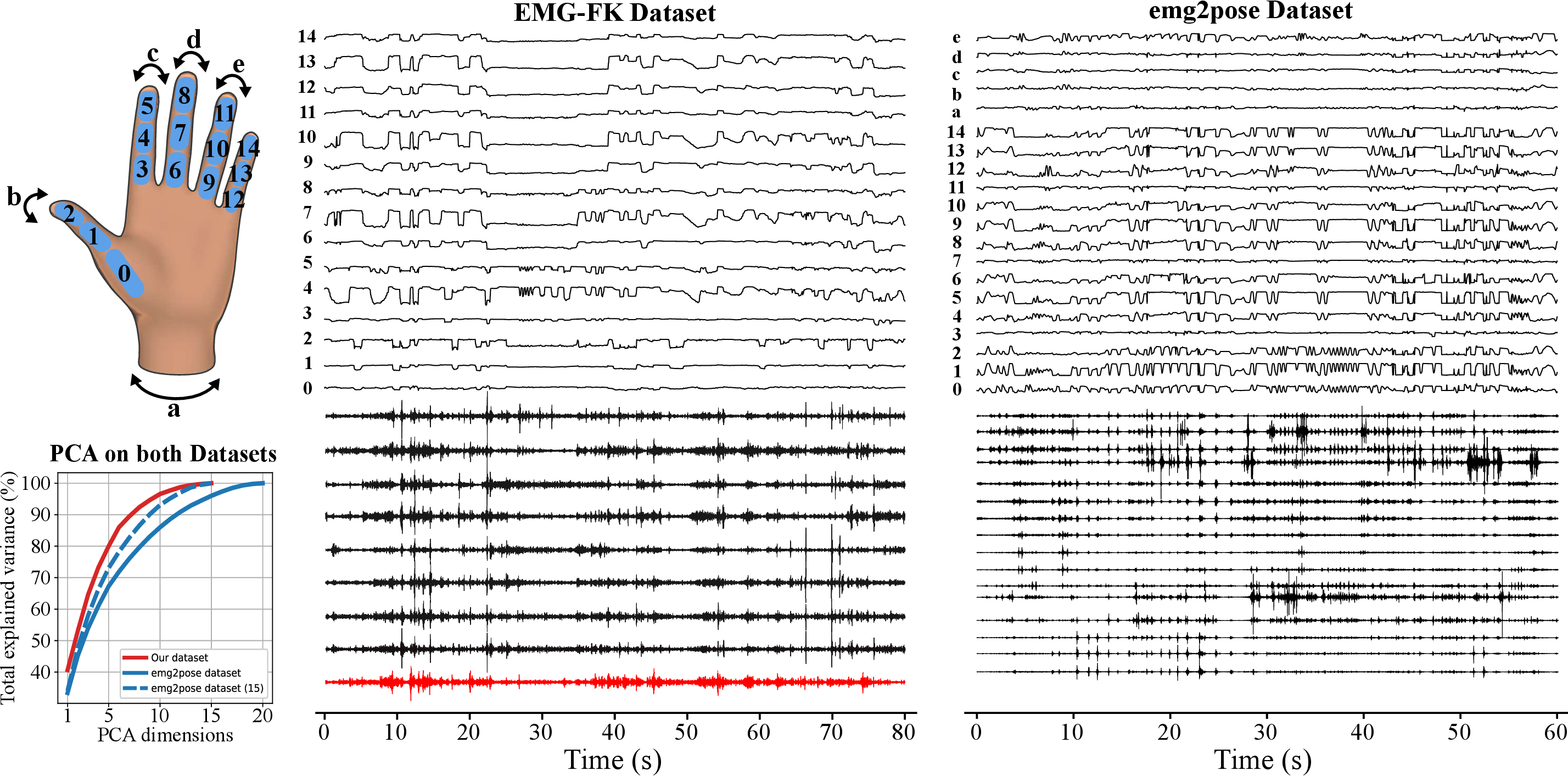}
    \caption{Comparison of the \ourData and \textit{emg2pose} datasets (joint-angles and EMG signals). The EMG from \ourData in this plot are processed using CAR. The EMG channel displayed in red represents the average value of all the recorded channels in \ourData before CAR.
    The bottom left plot shows the total explained variance ratio obtained with PCA on the joint angles of the two datasets, and on \textit{emg2pose} with only the 15 joint-angles that are present in both datasets.}
    \label{fig:datasetComparison}
\end{figure*}

We first perform data exploration to demonstrate that \ourData is suitable for joint-angle regression and comparable to the \textit{emg2pose} reference dataset. In Figure~\ref{fig:datasetComparison}, we illustrate the EMG signals and joint angles of both datasets, highlighting the diversity of finger gestures and muscular activations. 
Since our EMG sensor uses a single reference electrode rather than bipolar electrodes per channel, the signals across channels are strongly influenced by muscle activity at the reference location. To better highlight the diversity of individual muscle activations in \ourData, we apply common average referencing (CAR), i.e., subtracting the average across all channels at each time step. This preprocessing step improves data visualization. However, it is not used in the machine learning pipelines, as it may remove relevant information from the signal.

To analyze the diversity of gestures, we perform a principal component analysis (PCA) on the joint angles (Figure~\ref{fig:datasetComparison}). This shows that retaining only the first few principal components is insufficient to capture the full variance of the kinematics. Specifically, representing $90\%$ of the variance requires $7$ components for our dataset, $12$ for \textit{emg2pose}, and $9$ for a reduced version of \textit{emg2pose} that includes only the joint angles shared between both datasets.
These results suggest that \ourData exhibits only slightly less gestures variability  than \textit{emg2pose}. They also indicate that both datasets necessitate true regression models, as the kinematic signals cannot be efficiently reduced to discrete classes.

Finally, Table~\ref{tab:datasetComparison} provides a comprehensive overview of the characteristics of the two datasets used in this paper. For \textit{emg2pose}, we describe the subset of the full public dataset used to compute our regression results. In total, the complete dataset contains 193 participants, with an average of four sessions per subject. In our experiments, we use only one session from the first 30 participants.

\begin{table}
\centering
\setlength{\tabcolsep}{8pt} 
\caption{Datasets comparison}
\begin{tabular}{ l  c c }
\hline
 & \ourData & \textit{emg2pose} (subsampled)\\
\hline
\# Subjects & $20$ & $30$ (out of $193$)\\
Minutes/Subject & $30$ & $14.9 \pm 2.6$ \\
\# EMG sensors & $8$ & $16$ \\
\# Joint angles & $15$ & $20$ \\
Sampling Frequency & $500Hz$ & $2000Hz$ \\
Size & $2Gb$ & $\approx 10Gb$ \\
Tasks & free & guided + free \\
Sensor location & Forearm & Wrist \\
\hline
\end{tabular}
\label{tab:datasetComparison}
\end{table}

\subsection{Evaluation of \ourModel \label{sec:model_eval}}

\subsubsection{Benchmarking against state-of-the-art}
We conduct a benchmark evaluation by comparing \ourModel with several existing approaches. Our study covers both feature-engineering methods (e.g., TDF/CMTS combined with shallow or moderately deep regressors) and representation-learning approaches using minimally processed sEMG data (e.g., convolutional and deep LSTM networks). As discussed in Section~\ref{sec:intro}, much of the related work relies on deep convolutional and recurrent neural networks. Accordingly, we evaluate the deep CRNN architecture described in Section~\ref{sec:deepCRNN}, using either raw EMG signals or EMG envelopes as input.
Other approaches rely on TDF and MLP regressors. We therefore evaluate the TDF features defined in Section~\ref{sec:TDF} with the MLP architecture presented in Section~\ref{sec:mlp}. In addition, since CMTS features are widely used in EEG processing, we also include an MLP regressor using CMTS features in our comparison.
Finally, we assess the \textit{vemg2pose} model~\cite{metaEMG2pose}, which appears as the current state of the art for EMG-based regression on large-scale datasets.

\begin{figure}
    \centering
    \includegraphics[width=1\linewidth]{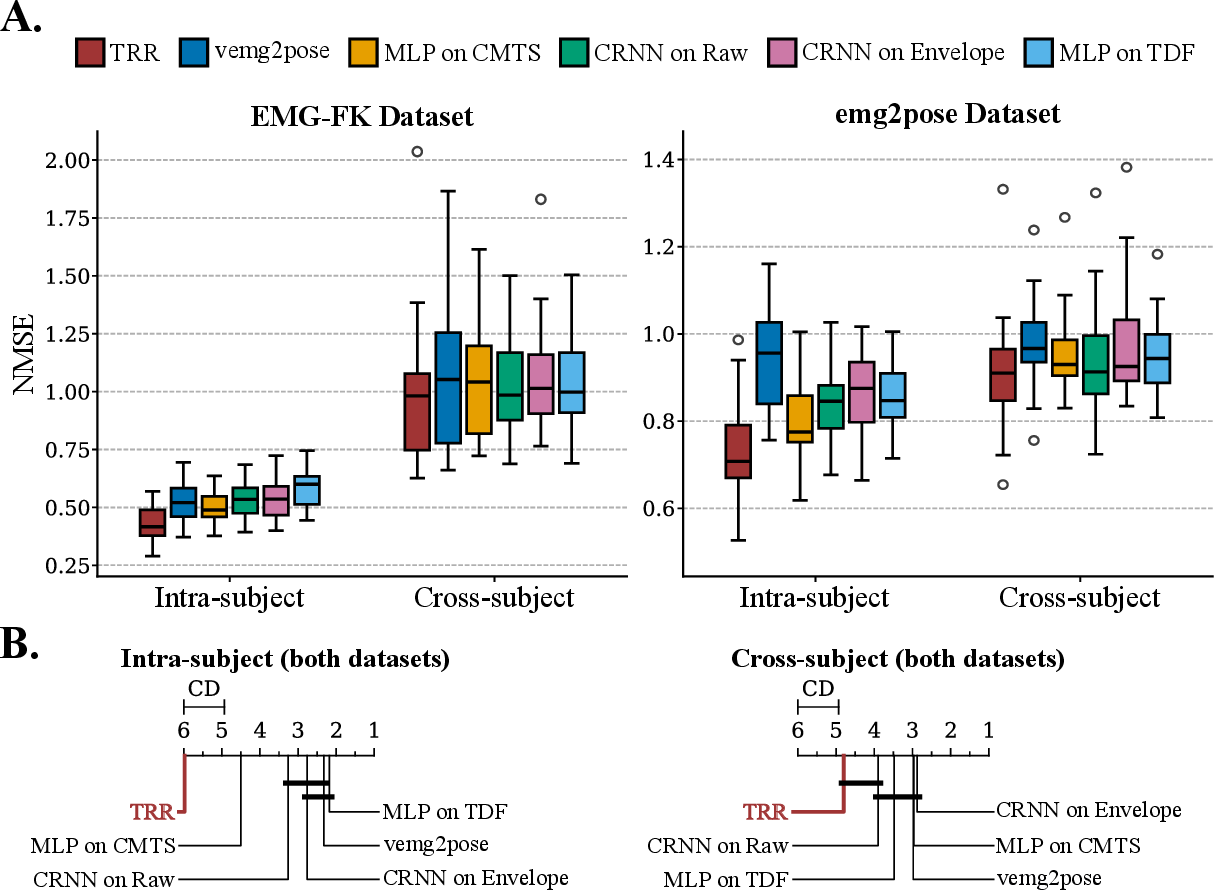}
    \caption{Evaluation of \ourModel on two datasets in comparison with state-of-the-art approaches, in intra- and cross-subject configurations. For statistical analysis, we compute the average rank of the models (higher is better). Models that are not significantly different are connected by a black line ($p = 0.05$, Nemenyi test~\cite{demvsar2006statistical}). The critical distance (CD) indicates when models are considered statistically different.}
    \label{fig:sota}
\end{figure}

In Figure~\ref{fig:sota}, we present the NMSE achieved by the different models on both the \ourData and \textit{emg2pose} datasets under intra-subject single-session and cross-subject configurations. The corresponding absolute errors on \ourData are reported in Table~\ref{tab:error_comparison}, as they offer more interpretable insights into model performance when expressed in degrees.

On \ourData, the proposed \ourModel model reaches an average intra-subject NMSE of $0.43 \pm 0.08$, corresponding to an average absolute (AE) error of $9.79\degree \pm 1.48$. On the \textit{emg2pose} dataset, it achieves an average intra-subject NMSE of $0.73 \pm 0.10$, while AE cannot be computed because the joint angles are scaled.
These results systematically outperform all baseline models, demonstrating that \ourModel, which was tuned using data from our acquisition framework, achieves significantly better intra-subject accuracy than existing approaches. In particular, CMTS feature vectors consistently provide the best support for regression, and our recurrent architecture further enhances gesture recognition.

\begin{table}
\centering
\setlength{\tabcolsep}{8pt} 
\caption{Absolute prediction error (in degrees) of the different methods on the \ourData dataset (mean $\pm$ std).}
\begin{tabular}{| l | c | c |}
\hline
\textbf{Baseline method} & \textbf{Intra-subject} & \textbf{Cross-subject} \\
\hline
\ourModel         & $9.79 \degree \pm 1.48$  & $16.71 \degree \pm 3.97$ \\
vemg2pose         & $11.04 \degree \pm 1.51$ & $16.76 \degree \pm 3.31$ \\
MLP on CMTS       & $11.41 \degree \pm 1.37$ & $18.66 \degree \pm 3.12$ \\
CRNN on Raw       & $11.35 \degree \pm 1.56$ & $17.87 \degree \pm 2.69$ \\
CRNN on Envelope  & $11.52 \degree \pm 1.52$ & $18.85 \degree \pm 2.76$ \\
MLP on TDF        & $12.93 \degree \pm 1.47$ & $18.70 \degree \pm 2.28$ \\
\hline
\end{tabular}
\label{tab:error_comparison}
\end{table}

Regarding cross-subject results, we observe that \ourModel also achieves the best overall score. However, the ranking is less pronounced, indicating that the best-performing method varies across test subjects.
Overall, cross-subject predictions are substantially less accurate than intra-subject predictions, particularly on the \ourData dataset, where the AE obtained by \ourModel goes from $9.79\degree \pm 1.48$ in intra-subject, to $16.71\degree \pm 3.97$ in cross-subject.

Interestingly, \textit{vemg2pose} exhibits higher error than all other methods on the \textit{emg2pose} dataset. An explanation can be that the characteristics of our experimental setup are different from the ones considered for tuning this model. In~\cite{metaEMG2pose}, a very large inter-subject, inter-session setting is used, with training data spanning over 300 hours of EMG. In contrast, our work focuses on lighter scenarios, where each model in the cross-validation is trained on less than 30 minutes of data for intra-subject, and about 15 hours of data for cross-subject configurations.
As \textit{vemg2pose} is inherently a deep learning approach, we can expect it to reach better results when trained on a very large quantity of data.

\subsubsection{Evaluation of \ourModel's gesture recognition}

\begin{figure*}
    \centering
    \includegraphics[width=1.0\linewidth]{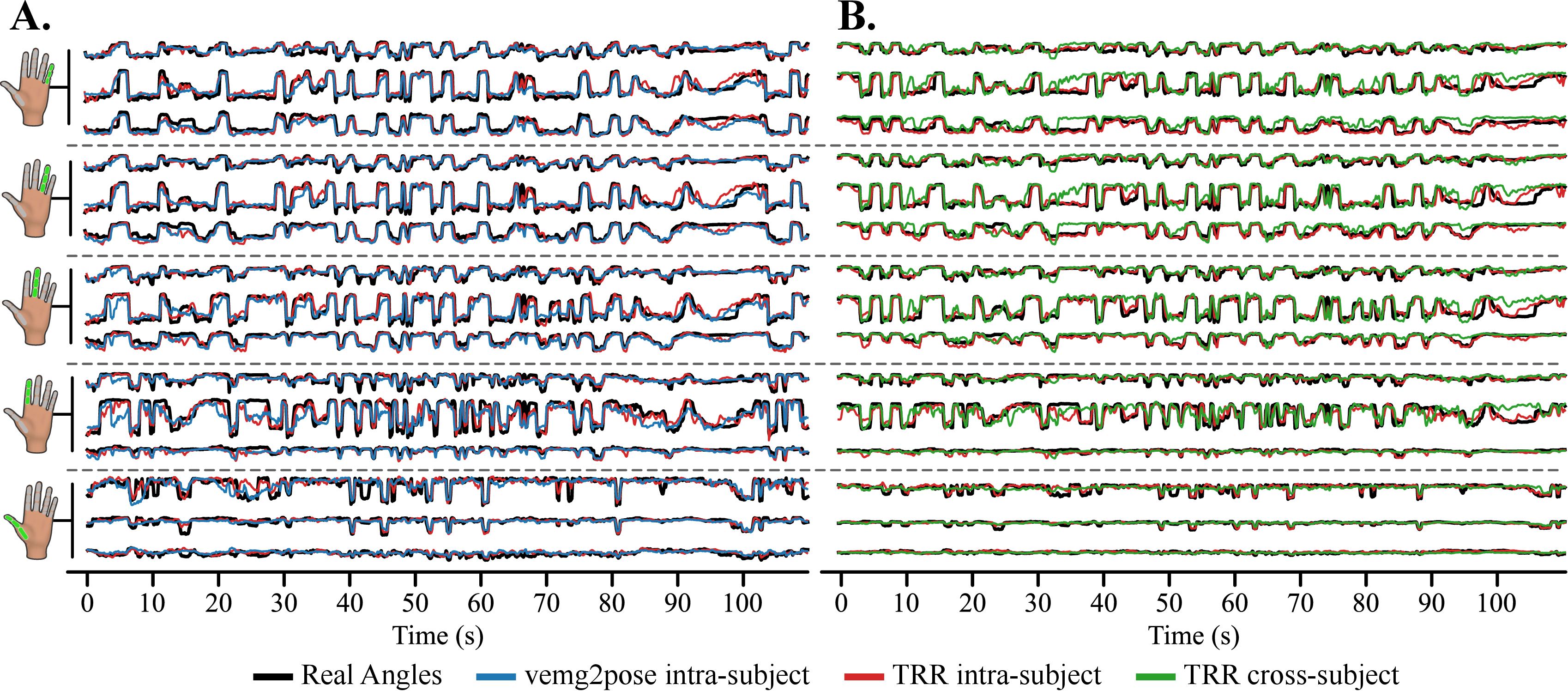}
    \caption{Visualization of the predicted joint-angles. \textbf{A.} Predictions from \ourModel and \textit{vemg2pose} on one subject from our \ourData dataset in intra-subject configuration. \textbf{B.} Predictions from \ourModel on one subject from \ourData in intra-subject and cross-subject configurations}
    \label{fig:anglesPlot}
\end{figure*}

In Figure~\ref{fig:anglesPlot}.A, we illustrate the intra-subject predicted kinematics for a participant from our dataset, comparing \ourModel with \textit{vemg2pose}. Although \ourModel achieves lower error overall, the predictions from both models often appear similar, with occasional instances where our approach outperforms the \textit{vemg2pose} baseline. 
In Figure~\ref{fig:anglesPlot}.B, we compare the predictions of \ourModel in intra-subject configuration with those from \ourModel in cross-subject configuration. In this setting, the difference is more important, but the overall shape of the cross-subject prediction still appears to match most of the real kinematics.
This observation suggests that, although there is still room for improvement of cross-subject regressors, \ourModel manages to capture relevant invariant information across subjects.
This behavior is further illustrated in a video reconstruction of the 3D hand model, showing side-by-side predictions from both models in both configurations~\footnote{Models comparison video: \url{https://www.youtube.com/watch?v=KrlFlxpkc6g}}.

\begin{figure}
    \centering
    \includegraphics[width=1\linewidth]{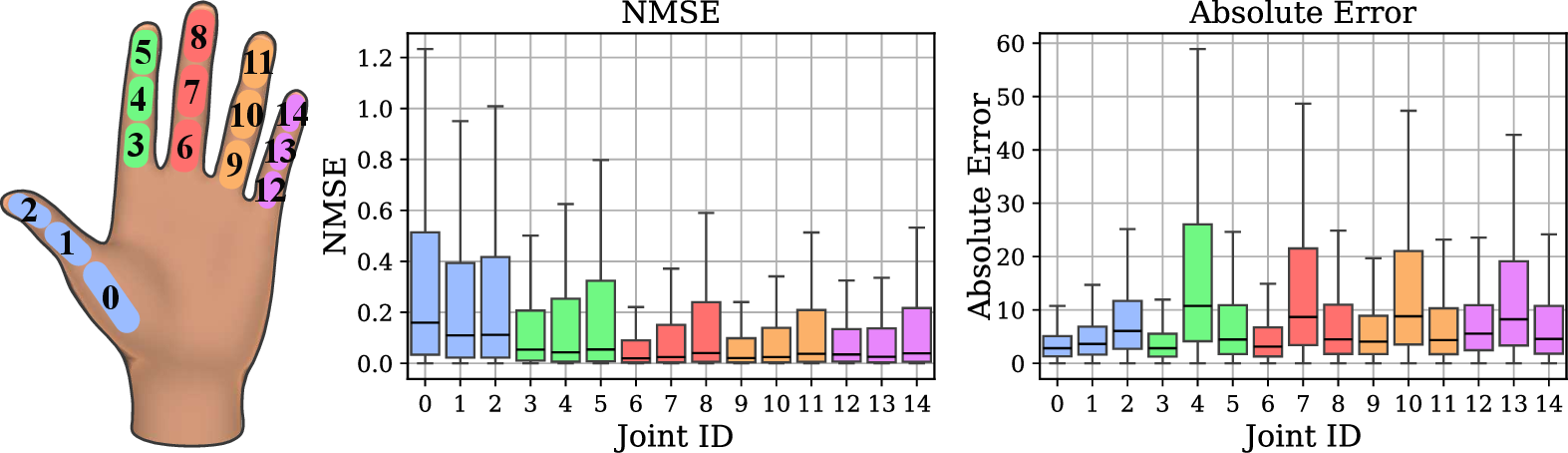}
    \caption{Evaluation of \ourModel on the different fingers joints, using all subjects from our \ourData dataset. The absolute error provides an estimation of the error angle in degrees.}
    \label{fig:evalPerFinger}
\end{figure}

In Figure~\ref{fig:evalPerFinger}, we report the intra-subject NMSE and absolute error of \ourModel for each joint. The results show that thumb gestures are less accurately recognized compared to the other fingers. This can be attributed to the thumb’s smaller range of motion, as captured by our motion tracking system, and the higher complexity of thumb movements relative to its anatomical capabilities. In contrast, gestures of the other fingers exhibit similar recognition accuracy.

\subsubsection{Ablation study \label{sec:ablation_study}}

\begin{figure*}
    \centering
    \includegraphics[width=\linewidth]{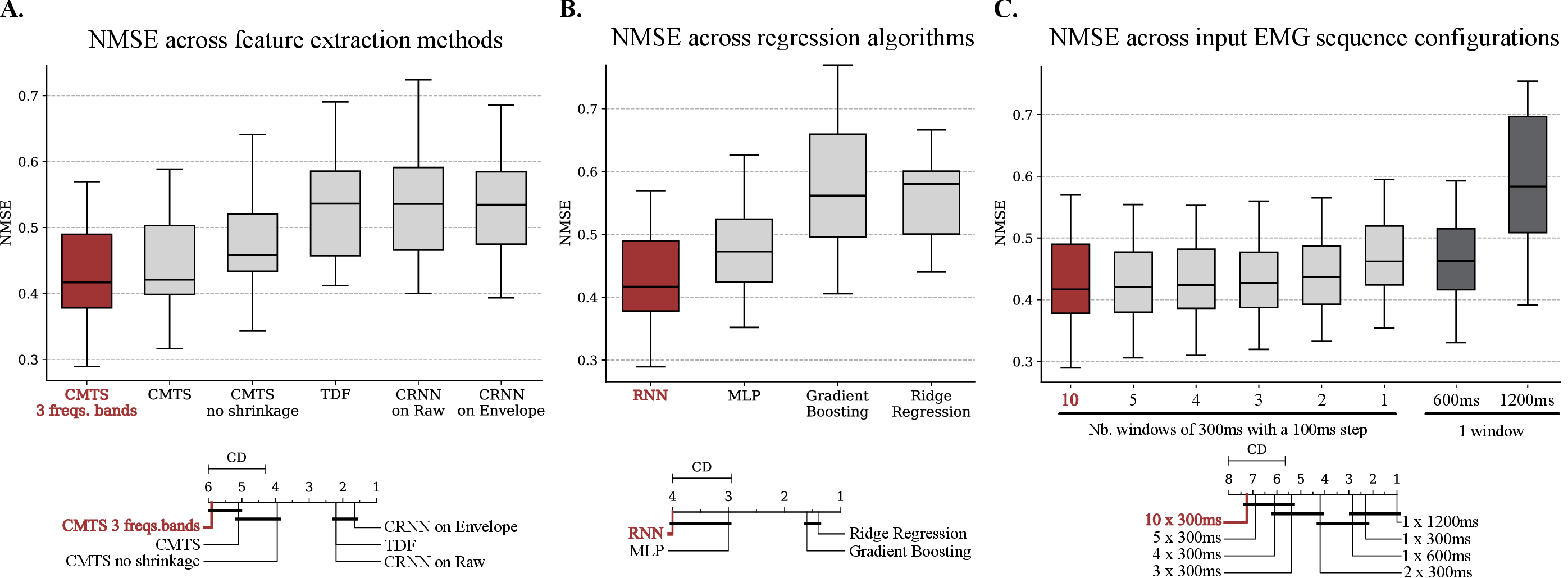}
    \caption{Results of the ablation study. The results in red correspond to \ourModel \textbf{A.} Effect of changing the feature extraction method while keeping our model architecture fixed. \textbf{B.} Effect of changing the regression algorithm using our chosen feature extraction method. \textbf{C.} Effect of varying the configuration of the input sequence for \ourModel. For statistical tests, we compute the average rank of the classifiers (the higher, the better). Those that are not significantly different are connected by a black line ($p = 0.05$, Nemenyi test~\cite{demvsar2006statistical}). The critical distance (CD) indicates when classifiers are considered statistically different.}
    \label{fig:ablationStudy}
\end{figure*}

In Figure~\ref{fig:ablationStudy}, we present the results of an intra-subject ablation study on our \ourData dataset, analyzing the impact of feature extraction methods, regression algorithms, and input sequence configurations to illustrate the design choices that optimize the \ourModel method.

First, for the feature extraction, we evaluate CMTS with three frequency bands (\ourModel), CMTS computed on a single full frequency band (15–150 Hz, simplified \ourModel), CMTS in the full frequency band without shrinkage, TDF, and deep convolutional features extracted from the raw EMG or the EMG envelope using the model described in Section~\ref{sec:deepCRNN}.
The results, reported in Figure~\ref{fig:ablationStudy}.A, show that \ourModel achieves the best performance and, in particular, that CMTS features significantly outperform both TDF and deep convolutional features.
Second, for the regression algorithm, we compare \ourModel (RNN-based) with an MLP, gradient boosting with regressor chains, and ridge regression. All models use the same sequence of CMTS feature vectors from three frequency bands as input. For models without recurrent architectures, the sequence is concatenated into a single input vector.
The results in Figure~\ref{fig:ablationStudy}.B show that \ourModel achieves the best performance. In particular, neural network–based models significantly outperform simpler regression approaches. This can be explained by the multivariate nature of the kinematic data and the non-linear relationships between joint-angles.
Third, to evaluate the effect of the input sequence configuration, we use the \ourModel's architecture and feature extraction method with different sequence lengths, as well as single input windows of larger sizes.
The results in Figure~\ref{fig:ablationStudy}.C show that shorter sequences and single longer input windows lead to higher errors. Moreover, performance converges when using a sequence of 10 windows, suggesting that this configuration provides sufficient temporal context for accurate prediction.

\subsubsection{Proof of concept: real-time robotic hand control}

In the previous sections, we reported that the intra-subject predicted motion closely follows the actual gestures. This suggests that \ourModel provides sufficiently accurate estimates for hand gesture recognition, enabling natural use in applications such as prosthesis control or AR/XR interfaces. To further demonstrate this, we showcase our method (\ourModel and \ourData acquisition framework) in a real-time control scenario with a robotic hand for one subject~\footnote{Robotic hand control demo video: \url{https://www.youtube.com/watch?v=fDK7dXqEAEI}}. While this demonstration is not a statistical evaluation, it illustrates that our approach allows intuitive control of a multi-degree-of-freedom robotic hand.

Furthermore, to validate that \ourModel can run on an embedded system, which is more realistic for commercial applications, we used a Raspberry Pi 5~\footnote{Raspberry Pi 5: \url{https://www.raspberrypi.com/products/raspberry-pi-5/}} to measure the average inference time per sample and the evolution of the CPU temperature over time.
The models are implemented in \textit{Python}. \textit{vemg2pose} uses \texttt{pytorch}~\cite{pytorch} for neural networks, and the other models use \texttt{tensorlow}~\cite{tensorflow2015}. The \texttt{pyriemann} package~\cite{pyriemann} is used to compute covariance matrices, and signal filtering is done using the \texttt{mne} library~\cite{mne}.
In the Table~\ref{tab:inferenceTime}, we show the average inference time per sample on the embedded system, excluding the initial band-pass filter, which is applied for all models. It appears that \ourModel is significantly faster than \textit{vemg2pose}, achieving almost 10 predictions per second in its simplified version (full frequency band), which is similar to the baseline \textit{MLP on TDF} approach.

\begin{table}
\centering
\setlength{\tabcolsep}{8pt} 
\caption{Average angle inference time with a RaspberryPi 5 on 500 samples from our dataset}
\begin{tabular}{ l | c c}
\hline
        & \multicolumn{2}{c}{Time per sample} \\
  Model & Feature extraction & Inference \\
\hline
\ourModel & $42 \pm 14ms$ & $105 \pm 23ms$ \\
\ourModel simplified & $1.5 \pm 0.5ms$ & $107 \pm 24ms$ \\
MLP on TDF & $0.7 \pm 0.2ms$ & $89 \pm 21ms$ \\
\textit{vemg2pose}  & $--$ & $919 \pm 42ms$ \\
\hline
\end{tabular}
\label{tab:inferenceTime}
\end{table}

\begin{figure}
    \centering
    \includegraphics[width=1.0\linewidth]{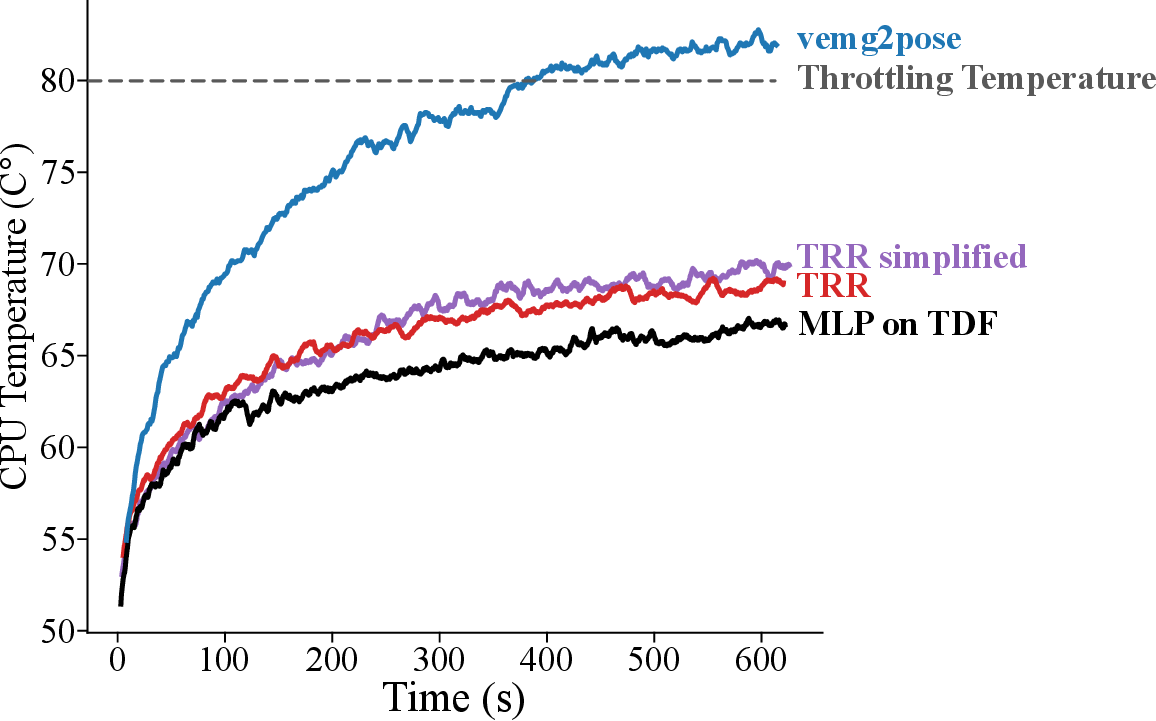}
    \caption{Evolution of the \textit{Raspberry Pi 5} CPU's temperature over time with different regression models including \ourModel and its simplified version (single frequency band), \textit{vemg2pose}, and a simple MLP on TDF to serve as baseline.}
    \label{fig:temperaturePlot}
\end{figure}

In Figure~\ref{fig:temperaturePlot}, we observe that the CPU temperature also rises substantially faster when using the \textit{vemg2pose} method, reaching the \textit{Raspberry Pi 5}'s throttling temperature in less than 10 minutes. This further highlights the potential advantages of our \ourModel model, which relies on engineered features rather than deep representation-learning approaches.

\subsubsection{Limitations toward myoelectric prosthesis control}

While the results of this study demonstrate that \ourModel outperforms state-of-the-art methods in predicting high-degree-of-freedom finger joint angles from sEMG and is better suited for embedded deployment, it is important to acknowledge limitations when generalizing these findings to applications in myoelectric prosthesis control.

First, our acquisition protocol was specifically designed to record continuous, rich finger kinematics during unconstrained motion while deliberately minimizing confounding factors such as grip-force generation, object manipulation, and concurrent arm or body movements (Section~\ref{sec:acquisitionProtocol}). As a result, the generalization of the learned mappings to real-world tasks involving variable external loads, fine motor manipulation, or artifacts from simultaneous upper-limb motions or body movements remains to be validated.

Second, because \ourModel is inherently lightweight compared to representation learning approaches, we did not explore quantization strategies. Such techniques could further reduce the memory footprint and enhance efficiency on edge devices with minimal accuracy degradation in EMG-based regressors~\cite{zanghieri2021}. Investigating quantized model variants could therefore expand the range of embedded platforms suitable for real-time inference. Moreover, quantized versions of deeper neural networks might become viable on certain embedded systems.

Finally, our current results are based exclusively on short single-session recordings. In practical scenarios, sEMG signal distributions drift over time due to electrode displacement, variations in skin-electrode impedance, and muscle fatigue. Thus, ensuring sustained performance, along with maintaining stability and safety in real-world conditions, will likely require periodic recalibration or the implementation of adaptive strategies such as continuous learning or domain adaptation.

\section{Conclusion}

In this work, we introduced a new open and cost-efficient framework for regression-based hand gesture recognition from EMG. This framework provides all the necessary components for EMG and joint-angle kinematics acquisition, synchronization, preprocessing, for feature extraction, and for model design.
We showed that this framework enables the collection of high-quality data, competitive with the \textit{emg2pose} dataset, which relies on a more complex and expensive acquisition procedure. Moreover, the recorded gestures involve complex and continuous, multi-degree-of-freedom hand movements that inherently require a regression approach, in contrast to many related studies where the problem can be simplified to classification.
We also presented \ourModel, a new regression model designed for real-time joint-angle prediction from EMG. Through multiple visualizations, we demonstrated that it achieves improved performance compared to existing methods in intra-subject and cross-subject settings. By relying on CMTS rather than deep feature extraction, \ourModel opens interesting perspectives for more interpretable EMG processing in human–computer interfaces.
Finally, by deploying \ourModel on a physical robotic hand, we demonstrated that our framework and model can be readily used in practical applications such as robotic prostheses.
Future work should explore two directions. First, aim to reduce the regression error in the more challenging cross-session and cross-subject scenarios by exploring domain adaptation methods. Second, study the practical use of the designed regression model in real-world applications, such as robotic hand control, which adds a lot of noise and context-specific effects to the EMG and kinematic signals.

\section*{Acknowledgments}
This publication benefits from the support of the Walloon Region to Pr. G. Bontempi as part of the funding for the FNRS‑WEL-T strategic axis. We would like to thank the research fund and the Leibu fund of the Université Libre de Bruxelles and of the Université de Mons, Belgium. We would like to thank Cléa Mafrica and Olalla Pruneda, and Robin Petit for their insightful feedback and unwavering support. Finally, we gratefully acknowledge the support of the FARI Institute and TRusted AI Labs for their continued commitment to advancing research and innovation. We acknowledge the use of ChatGPT (OpenAI) solely for improving the clarity and quality of the writing; it did not contribute to the scientific content, analysis, or results of this work.

\section*{Data and Code Availability}
The acquisition framework, Python implementation of \ourModel, and benchmark evaluation code are publicly available on GitHub at~\url{https://tinyurl.com/2j8y3a3n}. The \ourData{} dataset is publicly available on the Zenodo repository~\cite{EMGFK_dataset}.

\bibliographystyle{IEEEtran}
\bibliography{biblio}

\clearpage

\appendices

\section{Impact of the duration of the training dataset}
\label{sec:evalDurationTrainingData}

To evaluate the quantity of training data necessary to obtain optimal intra-subject error for joint-angle regression with \ourModel, we perform multiple cross-validations on the \ourData dataset by randomly selecting a subset of training samples (in the form of a continuous sequence of samples).

\begin{figure}
	\centering
	\includegraphics[width=1\linewidth]{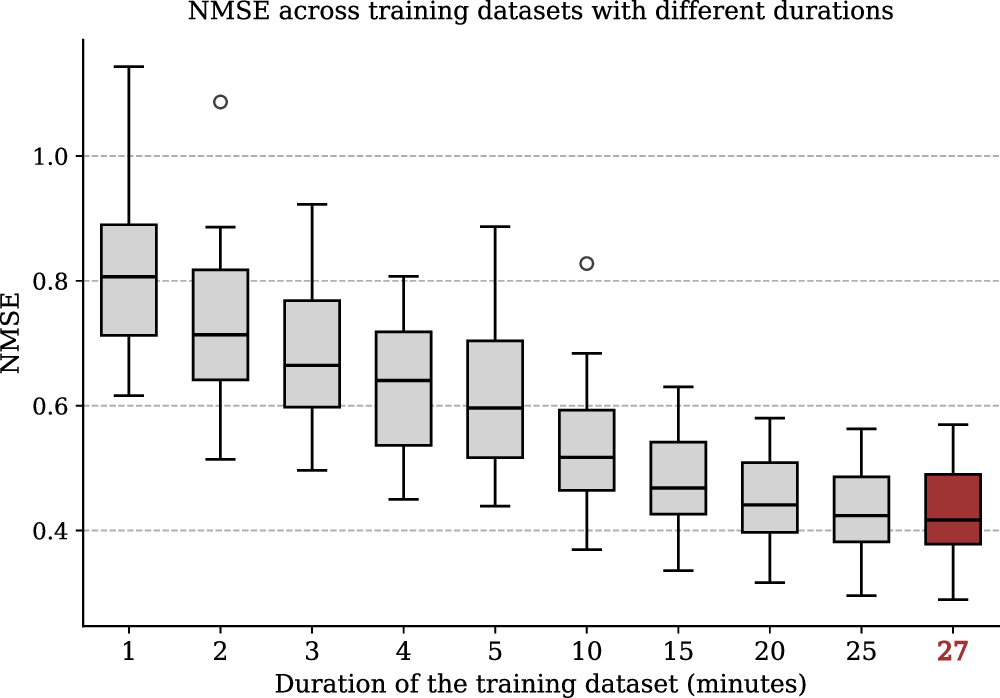}
	\caption{Evaluation of \ourModel with different durations of training data}
	\label{fig:impactDurationTrainingData}
\end{figure}

In Figure~\ref{fig:impactDurationTrainingData}, we report the intra-subject NMSE for different duration of the subsampled training data. For each duration, 10\% of the training data is kept aside for validation to determine training convergence and avoid overfitting.
We see that the error only begin to converge when using all the available training dataset. This suggest that \ourModel needs to leverage large quantity of data in order to obtain the best regression results even within a single subject.

\section{Evaluation of the synchronization procedure}
\label{sec:evalSync}

To evaluate our synchronization procedure, we perform cross-validation after applying different temporal offsets between the EMG and kinematics after synchronization.
If the synchronization procedure is perfect, when the offset is positive, the model should perform forecasting of the gestures for which the EMG has not yet been produced. We thus expect a significant drop in performance.
On the other hand, with negative offset, delayed in introduced in the prediction, as the model now has access to the EMG produced after the target motion. We expect a slight improvement in performance here.

\begin{figure}[H]
	\centering
	\includegraphics[width=1\linewidth]{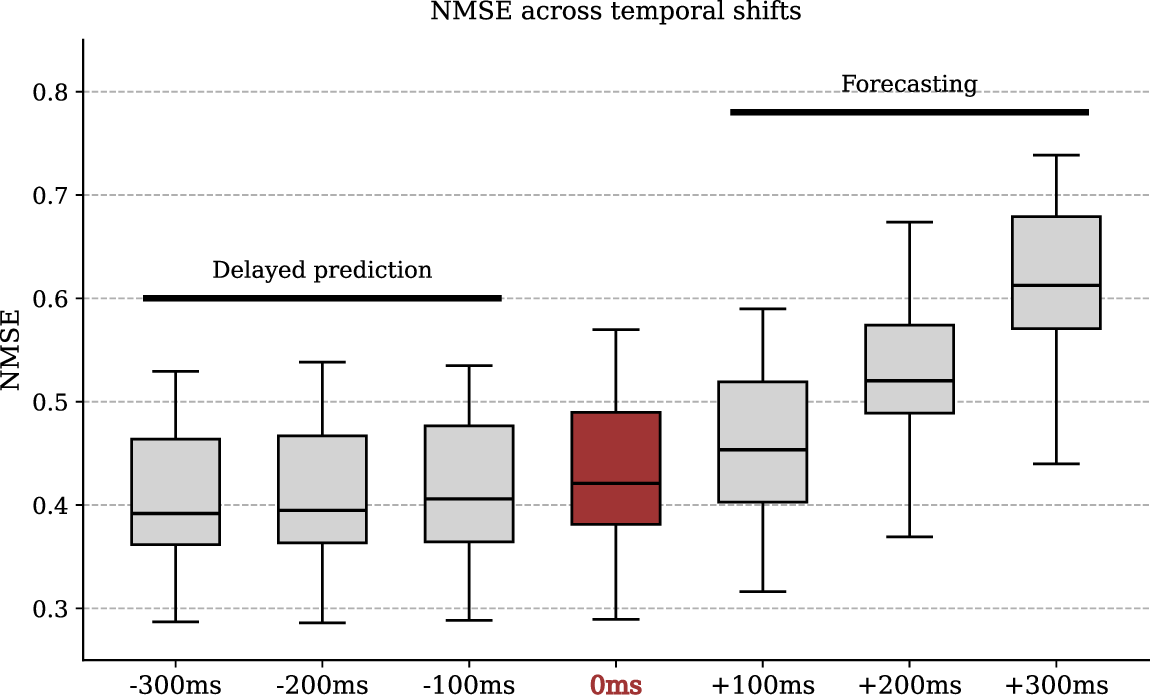}
	\caption{Evaluation of \ourModel with different temporal offset between EMG and joint-angle kinematics}
	\label{fig:impactTemporalShift}
\end{figure}

The Figure~\ref{fig:impactTemporalShift}, reports the results on the \ourData dataset, confirming the accuracy of our synchronization method.
We must note that this approach does not take into account the physiological delay that can occur between the production of EMG and the actual hand gesture. However, it seems that the gain in performance obtained when introducing delay in the predictions is minimal. From a practical point of view, it is thus more interesting to provide kinematic predictions as fast as possible.

\end{document}